# ICPS-net: An End-to-End RGB-based Indoor Camera Positioning System using deep convolutional neural networks

Ali Ghofrani[1], Rahil Mahdian Toroghi[1], and Sayed Mojtaba Tabatabaie[2]

[1] Faculty of Media Technology and Engineering, Iran Broadcasting University, Tehran, Iran
[2] CEO/CTO at Alpha Reality, AR/VR Solution Company
Emails: alighofrani@iribu.ac.ir, mahdian.t.r@gmail.com, smtabatabaie@alphareality.io

## ABSTRACT

Indoor positioning and navigation inside an area with no GPS-data availability is a challenging problem. There are applications such as augmented reality, autonomous driving, navigation of drones inside tunnels, in which indoor positioning gets crucial. In this paper, a tandem architecture of deep network-based systems, for the first time to our knowledge, is developed to address this problem. This structure is trained on the scene images being obtained through scanning of the desired area segments using photogrammetry. A CNN structure based on EfficientNet is trained as a classifier of the scenes, followed by a MobileNet CNN structure which is trained to perform as a regressor. The proposed system achieves amazingly fine precisions for both Cartesian position and quaternion information of the camera.

## 1. INTRODUCTION

Global positioning system is still a universal challenging problem, which has been contributed using navigation systems, and GPS satellites. The precision of positioning and the ubiquitous usability of such systems, especially around the tall buildings or covered areas, are remaining unsolved, though. The indoor positioning would be also used when the GPS systems are not accessible, such as inside tunnels, malls, huge buildings, airports, skyscrapers, and so on. While for open areas using GPS system is the most prevalent solution, for covered places the conventional methods mostly incorporated the image processing techniques such as SIFT and SURF, for which the results were not so accurate and their precision were not satisfactory, at all. The main reason was the existence of several identical and repetitive patterns inside the buildings, which could easily hinder the uniqueness of the positions.

The first deep network-based solution for positioning, were introduced in ICCV 2015 [1], however for a limited open area. The system called POSENET, and consists of a convolutional neural network for a real-time 6 degree-of-freedom camera relocalization. Along the way, a geometry-aware system was introduced to the maps for camera localization which incorporated perceptual and temporal features to improve the precision, [2]. Both of these methods, could address the outdoor positioning problem. For indoor positioning, the conventional methods either used a depth-assisted RGB camera [3], or standard SIFT-based algorithm [4], if image processing were involved. Otherwise, WiFi and Bluetooth signals were used [5]. In real world scenarios, the depth-assisted camera is not always available. Moreover, we limit ourselves to the condition in which no WiFi and Bluetooth signals could work, since these methods are not so accurate. There is still a tangible lack of techniques to be employed for indoor localization and mapping purposes. The difficulties are about, 1) the accuracy of the position estimates and the camera angle, as well as its point of view, 2) the drift problem of vSLAM (video-based Simultaneous Localization and Mapping) system, and 3) inaccessibility of GPS data.

In this paper, we are going to develop an end-to-end system based on image processing techniques, along with deep convolutional neural networks to perform the accurate indoor positioning without the aforementioned problems. Besides, a new dataset is released which could be used for training the neural networks, as well as evaluation and comparison of different techniques within the scope of the indoor positioning problem. This dataset would be freely available for research purposes [6].

The outline of the paper appears as the following: First, the structure of the system is introduced. Then, the details of the experiments, the setup condition and the evaluation results are explained. The conclusion, wraps up the entire work in few fundamental statements, and the paper is finalized by the references appeared sequentially in the paper.

## 2. INDOOR POSITIONING SYSTEM STRUCTURE

In order to perform the positioning, the following steps should take place in a sequence: 1) The place, should be divided into scenes with possibly different patterns, 2) From every scene several image samples should be taken with a sufficient variety of camera movements and view point styles, and these sequence of samples are to be saved in a dataset, 3) A deep neural network system is to be trained to perform the scene classification, 4) Another deep neural network system is to be trained to perform the positioning and output the location information as well as the quaternion. This network, performs the position regression, indeed. A complete training and testing work-flow is depicted in figure 1. Training of these networks are performed across all the layers.

The complete functional diagram of the training and testing phase for the proposed indoor positioning system (ICPS-net). During the training the scene is selected, then the sequence of images from the selected scene is created through photogrammetry, and appended to the database. A selected CNN is then trained to output the XZY-positions as well as the Quaternion (WPQR). During the Test (the bottom image), an image is fed to a classifier network, (e.g., an EfficientNet, ShuffleNet V2 or MobileNet V2) to estimate the scene first, and then the regressor CNN will estimate the positions and Quaternion (WPQR) information, as it has already been trained for.

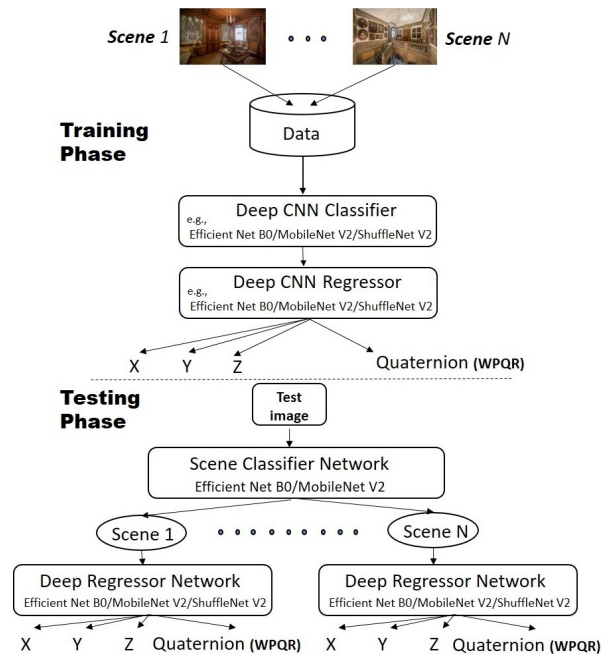

Figure 1. Functional diagram of the training and testing phase

As mentioned above, prior to training the included scenes inside the positioning area are turned into several sample images based upon some pre-defined camera trajectory styles. Some of these styles are depicted in figure 2. These styles are: rectangular, spiral, circular, semicircular, and random-wise with forward and backward camera viewpoints. The scenes are sampled based on the above camera trajectory styles, and the output sequence of the images, as the result of the camera movement, are saved inside the dataset as the sequence of images associated to each scene.

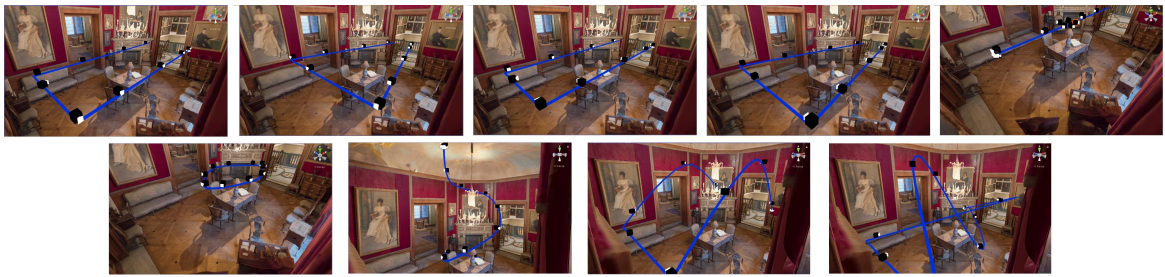

Figure 2. Sample camera movement trajectories

The test dataflow of the system, is as follows: 1) The user feeds the test image through the scene classifier network, depicted in figure 3. This network gets the input image, performs some preprocessing and then passes the image of size (224*224*3) to the Efficient Net B0 network [7]. The output of this network, after multiple dense layers is a softmax, which outputs the most likely scene number. 2) When the most likely scene is estimated, the output should be passed through the second network, which is depicted in figure 4. Further, this network gets the image, which is supposed to belong to a specific scene, and estimates the position (X,Y,Z) and quaternion information (wpqr) using a structure.

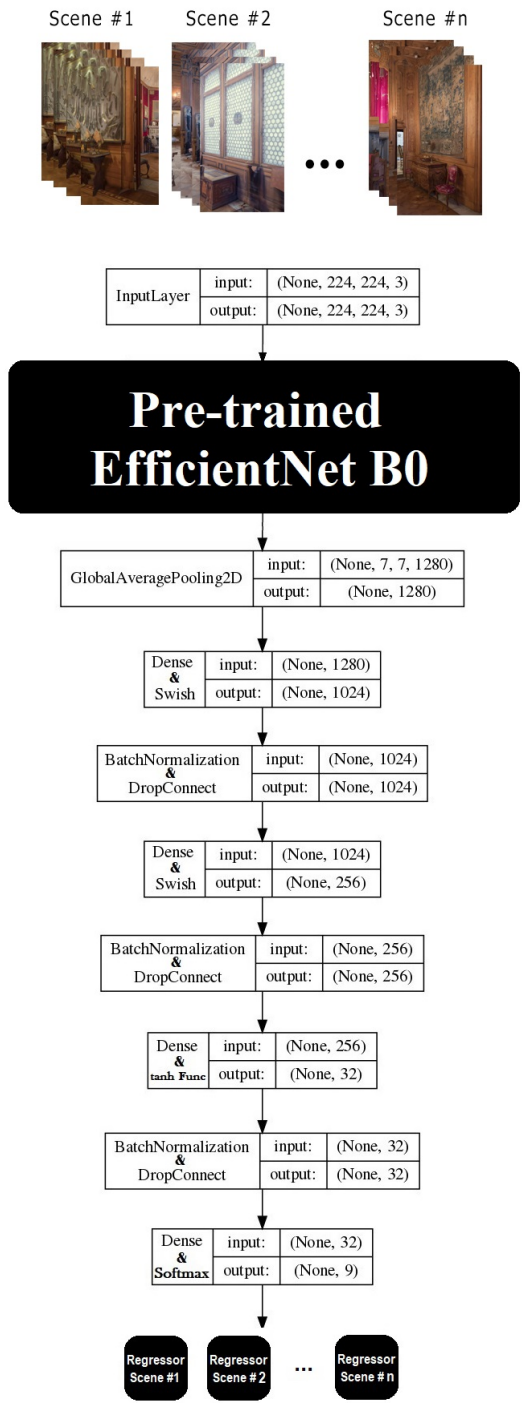

Figure 3. The classifier deep network architecture

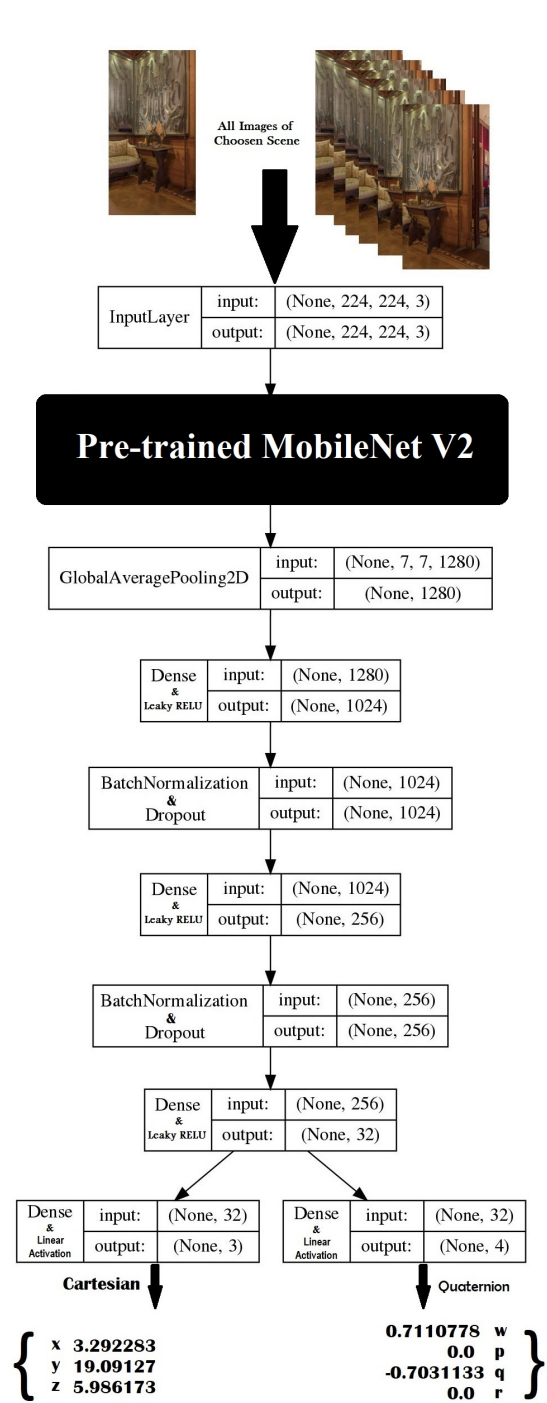

Figure 4. The regressor deep network architecture

The generated images associated to each scene assist the network to localize any point inside the scene, due to the variety of camera points of view and movement styles. A universal origin point has been chosen for all the scenes. This helps us train a unified neural network for each scene, or a set of scenes, or all the scenes together.

One important point is the impossibility of augmenting the data using the conventional methods like zooming, shear or rotating. The reason is that, these operations will change the position and quaternion and will directly affect the outputs. The only permissible augmentations are adding noise to the images, changing the brightness, an RGB channel shifting. The post-processes performed in between the dense layers are batch-normalization and drop out to avoid overfitting of the network.

The Euclidean loss function is employed for training the CNN regressors, which uses the Adam optimizer to minimize the loss [8], and is shown as follows [9],

$$loss = ||P - \hat{P}||_2 + \beta \, ||\hat{Q} - \frac{Q}{||Q||}||_2 \qquad (1)$$

where P = [x, y, z] is the position data vector, Q is the quaternion information, and β is the scale factor to make a balance between estimating the position and the quaternion. Moreover, every dense layer in the classifer network is ended with a swish function [10]. This function looks pretty much similar to the ReLU activation function, however it is smooth, non-monotonic, and as it is mentioned by Ramachandran et al. [10], Swish consistently matches or outperforms ReLU on deep networks applied to a variety of challenging domains.

## 3. EXPERIMENTS AND ANALYTICS

Primarily, the dataset has been created, based on the freely available 3D scanned images of the Hallwyl museum in Stockholm [11]. The first floor of this museum is scanned using photogrammetry. The 3D model is then loaded into the Unity-3D software. Using this software, we have sampled from each of the scenes with different camera movement styles and view angles which has been depicted in figure 2. For each photo, a 7- elements vector has been created, including the Cartesian positions plus the quaternion information of the camera (i.e. wpqr).

Table 1. Scenes and number of data samples being created for each, based-on cameras trajectories and view-angles

| Scene \ Seq | Seq1 | Seq2 | Seq3 | Seq4 | Seq5 | Seq6 | Seq7 | Seq8 | Seq9 | All |
|---|---|---|---|---|---|---|---|---|---|---|
| Armoury | 1739 | 1414 | 3081 | 1697 | 1213 | 1238 | 2064 | 2030 | 3113 | 17589 |
| Billiard | 1868 | 2229 | 2061 | 1735 | 1856 | 1851 | 2066 | 1495 | 1864 | 17025 |
| Dinning | 2229 | 2003 | 1767 | 1964 | 1694 | 1956 | 1983 | 1682 | 1813 | 17091 |
| Great Drawing | 1994 | 1581 | 2016 | 1713 | 1731 | 1926 | 1729 | 1705 | 2049 | 16444 |
| Morning | 2229 | 1925 | 2036 | 2148 | 1767 | 1917 | 2030 | 1615 | 2046 | 17713 |
| Porcelain | 2256 | 1708 | 1903 | 2021 | 1821 | 1503 | 1955 | 1618 | 1754 | 16539 |
| Serving | 2568 | 2002 | 2610 | 2158 | 2185 | 2390 | 2387 | 2034 | 2260 | 20594 |
| Small Drawnig | 2167 | 1702 | 1966 | 1979 | 1912 | 2012 | 1870 | 1442 | 1854 | 16904 |
| Smoking | 1745 | 1709 | 2016 | 2009 | 2004 | 1953 | 1990 | 1601 | 2023 | 17050 |
| All | 18795 | 16273 | 19456 | 17424 | 16183 | 16746 | 18074 | 15222 | 18776 | 156949 |

The scenes are divided into the following: Billiard, Great Drawing, Porcelain, Smoking, Dining, Morning, Serving, Armoury, and Small Drawing rooms. For the scenes, 9 different movement styles are defined which create 9 different sequences of sampled images, including: Seq1 X-Y forward rectangular, Seq2 X-Y backward rectangular, Seq3 X-Y

Forward Trapezoid, Seq4 X-Y Backward Trapezoid, Seq5 XY Straight Rotate, Seq6 X-Y Central Circle Rotate, Seq7 Z-Spiral, Seq8 Z Semicircular, and Seq9 Fully randomized. Table 1, shows the number of data samples being created from sampling the scenes using the above sequences. The output position data has been normalized based on the following,

$$P = 2\frac{P - Min}{Max - Min} - 1 \qquad (2)$$

where *Max* and *Min* are the maximum and minimum samples of each class, respectively. The created data has been divided into 60% training, 20% validation, and 20% test data samples. The training and validation accuracy and loss curves for the classifier are depicted in figure 5, and figure 6, respectively. In addition, the regressor training and validation loss, as well as the test results for regressor outputs over 12000 samples are depicted as a confusion matrix in figures 7, and figure 8, respectively. The final error values, achieved based on the proposed structures of figure 3, and 4, on the entire test data that the model has not seen before, is depicted in table 2.

Table 2. The average regression error (over meter), for the position vector (X;Y;Z), and the average camera Quaternion (over degree)

|  | X-Position | Y- Position | Z- Position | Quaternion |
| --- | --- | --- | --- | --- |
| Error Value | 0.0026 m | 0.0010 m | 0.0034 m | 0.0086 ° |

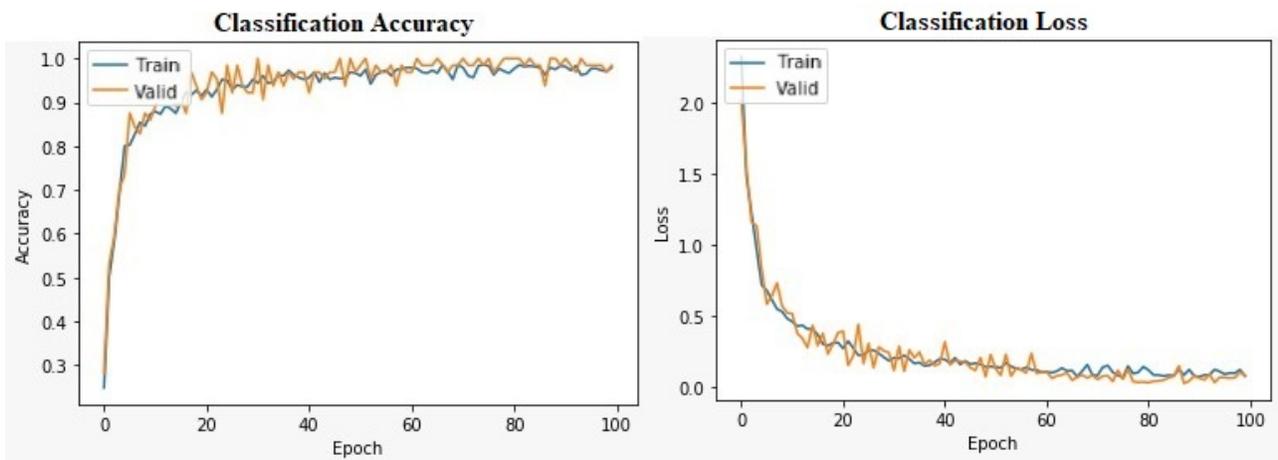

Figure 5. Classifier accuracyFigure 6. Classifier loss

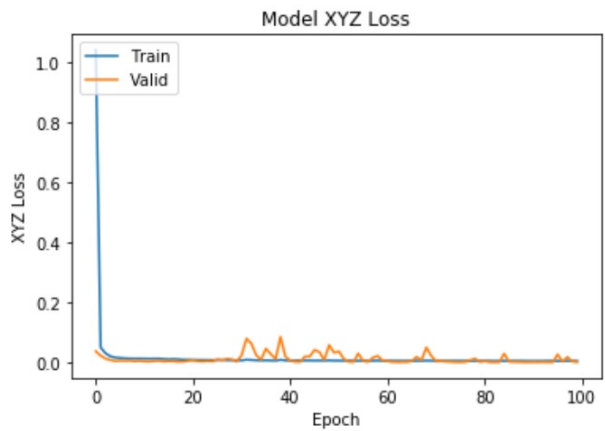

Figure 7. The regressor training and validation loss

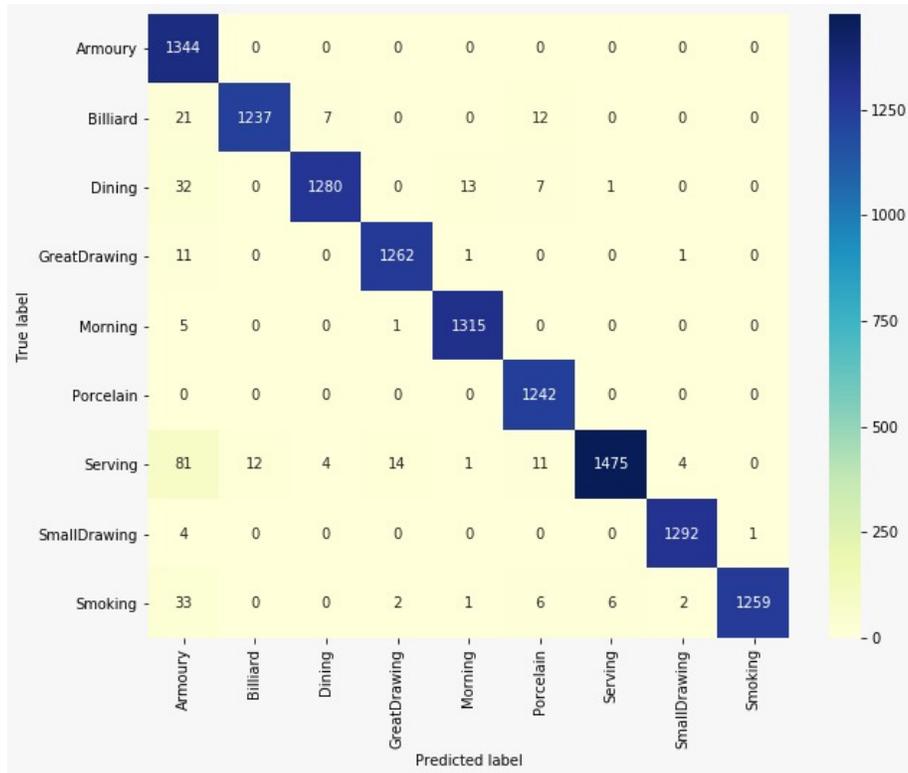

Figure 8. The confusion matrix over 12000 randomly chosen test samples

The EfficientNet-B0 contains 4,050,000 parameters, therefore it has been employed in the classification task. Since, we are authorized to perform augmentation in the classification structure, we could easily train it without the fear of overfitting. However, for the regressors since the data samples which could be used for training is limited to the number of samples belonging to the same class, and further we are not allowed to perform augmentation, therefore we employed a different model including less number of parameters, here specifically we incorporated the MobileNet V2 [12], containing about 2,250,000 parameters. To avoid overfitting problem, all the necessary precautions are considered, such as drop out, and drop-connect [13] and penalizing the activation functions, as well as batch normalization.

The loss function being involved in classification task is the categorical cross entropy. The network attempts to maximize the validation accuracy and is monitored persistently until it reaches the 100% validation accuracy. The validation loss, becomes 0.021851, the train accuracy reaches 98.514, and the training loss reaches 0.06415. After training is accomplished, we evaluated the model using the test data, and we could reach the accuracy of 98.099, and the loss of 0.063899.

The hardware system being used for this purpose, is a CUDA-core support GTX 1080, using a CPU of intel7700 core i7, and 16GB of RAM, which limits us from using large mini-batch sizes. The framework is Tensorflow 1.13.1, with Keras 2.2.4.

## 4. CONCLUSION

In this paper, the indoor positioning problem based on a supervised deep network structure has been addressed. The goal of the system is to achieve a high accuracy of the Cartesian (X,Y,Z) position and the camera quaternion. Using a tandem structure of Convolutional neural networks and dense (i.e. MLP) layers, one as a scene classifier (EffcientNet B0) and another as a position regressor (MobileNetV2), we could develop a system, called ICPS-net, with which we could achieve a very high position accuracy rate. The required dataset has been created through a laborious endeavor of sampling images from a 3D scanned photogrammetry of a Museum, which is freely available. In future, there should be serious attempts on how to create the required data in a simpler fashion, to make the training and inference tasks fast and feasible. The proposed deep network, could be used as the initializer of the customized projects in real world problems using transfer learning.

# ACKNOWLEDGEMENT

This work is based on the research supported wholly by Alpha Reality Company, we also owe to Mr. Seyed Maziar Tabasi, the PhD student of the University of Tehran, for his friendly support during creation of the dataset, by taking the samples from the 3D model of the Hallwyl museum.

# AUTHORS' BACKGROUND

| Your Name | Title | Research Field | Personal website |
| --- | --- | --- | --- |
| Ali Ghofrani | Master Student | Deep Learning, Computer Vision | ghofrani.ir |
| Rahil Mahdian Toroghi | Assistant professor | Machine Learning, Deep Learning | Scholar link |
| Mojtaba Tabatabaie | AR/VR Researcher | Augmented and Virtual Reality | alphareality.io |